\documentclass[11pt,a4paper]{article}
\usepackage[hyperref]{emnlp-ijcnlp-2019}
\usepackage{times}
\usepackage{latexsym}
\usepackage{soul}

\usepackage{url}

\usepackage{graphicx}
\usepackage{subcaption}
\usepackage[utf8]{inputenc}
\usepackage[export]{adjustbox}
\usepackage{wrapfig}
\usepackage{tabularx}
\usepackage{array,booktabs}
\usepackage{amssymb}
\usepackage{amsmath}
\usepackage{dsfont}
\usepackage{booktabs}

\aclfinalcopy 

\newcommand{\fromZ}[1]{\textcolor{violet}{#1}}
\newcommand{\fromH}[1]{\textcolor{blue}{[Huan] #1}}
\newcommand{\fromY}[1]{\textcolor{orange}{[Yu] #1}}

\newcommand{\nop}[1]{}

\newenvironment{remark}[1][Remark]{\begin{trivlist}
\item[\hskip \labelsep {\bfseries #1}]}{\end{trivlist}}

\title{Model-based Interactive Semantic Parsing:\\A Unified Framework and A Text-to-SQL Case Study}

\author{
    Ziyu Yao\textsuperscript{1}, Yu Su\textsuperscript{1}, Huan Sun\textsuperscript{1}, Wen-tau Yih\textsuperscript{2}\thanks{\hspace{.07in}Work started while at AI2}\\
    {\tt \{yao.470, su.809, sun.397\}@osu.edu}\\
    {\tt scottyih@fb.com}\\
    \textsuperscript{1}The Ohio State University\\
    \textsuperscript{2}Facebook AI Research, Seattle\\
}

\date{}

\begin{document}
\maketitle
\begin{abstract}

As a promising paradigm, \emph{interactive semantic parsing} has shown to improve both semantic parsing accuracy and user confidence in the results.
In this paper, we propose a new, unified formulation of the interactive semantic parsing problem, where the goal is to design a \textit{model-based} intelligent agent. The agent maintains its own state as the current predicted semantic parse, decides whether and where human intervention is needed, and generates a clarification question in natural language. A key part of the agent is a world model: it takes a percept (either an initial question or subsequent feedback from the user) and transitions to a new state. We then propose a simple yet remarkably effective instantiation of our framework, demonstrated on two text-to-SQL datasets (WikiSQL and Spider) with different state-of-the-art base semantic parsers.  Compared to an existing interactive semantic parsing approach that treats the base parser as a black box, our approach solicits less user feedback but yields higher run-time accuracy.\footnote{Code available at \url{https://github.com/sunlab-osu/MISP}.}

\end{abstract}

\section{Introduction}

Natural language interfaces that allow users to query data and invoke services without programming have
been identified as a key application of semantic parsing~\cite{berant2013semantic, thomason2015learning, dong2016language, zhong2017seq2sql, campagna2017almond, su2017building}.  However, existing semantic parsing technologies often fall short when deployed in practice,
facing several challenges: (1) user utterances can be inherently ambiguous or vague, making it difficult to get the correct result in one shot, (2) the accuracy of state-of-the-art semantic parsers are still not high enough for real use\nop{ (20\% to 90\% depending on the task and dataset)}, and (3) it is hard for users to validate the semantic parsing results, especially with mainstream neural network models that are known for the lack of interpretability.

\nop{
Semantic parsing has been proven to be a promising way for building natural language interfaces that allow users to
query data and invoke services without programming \cite{berant2013semantic, thomason2015learning, dong2016language, zhong2017seq2sql, campagna2017almond}. However, it has not seen wide applications in practice mainly due to three reasons: (1) user utterances are often inherently ambiguous or vague, making it impossible to get the perfectly right result in one shot, (2) state-of-the-art semantic parsers still often fall short in accuracy for real use\nop{ (20\% to 90\% depending on the task and dataset)}, and (3) it is hard for users to validate the parsing results, especially with mainstream neural network models that are known for lack of interpretability.
}

\begin{figure}
    \centering
    \includegraphics[width=.85\linewidth]{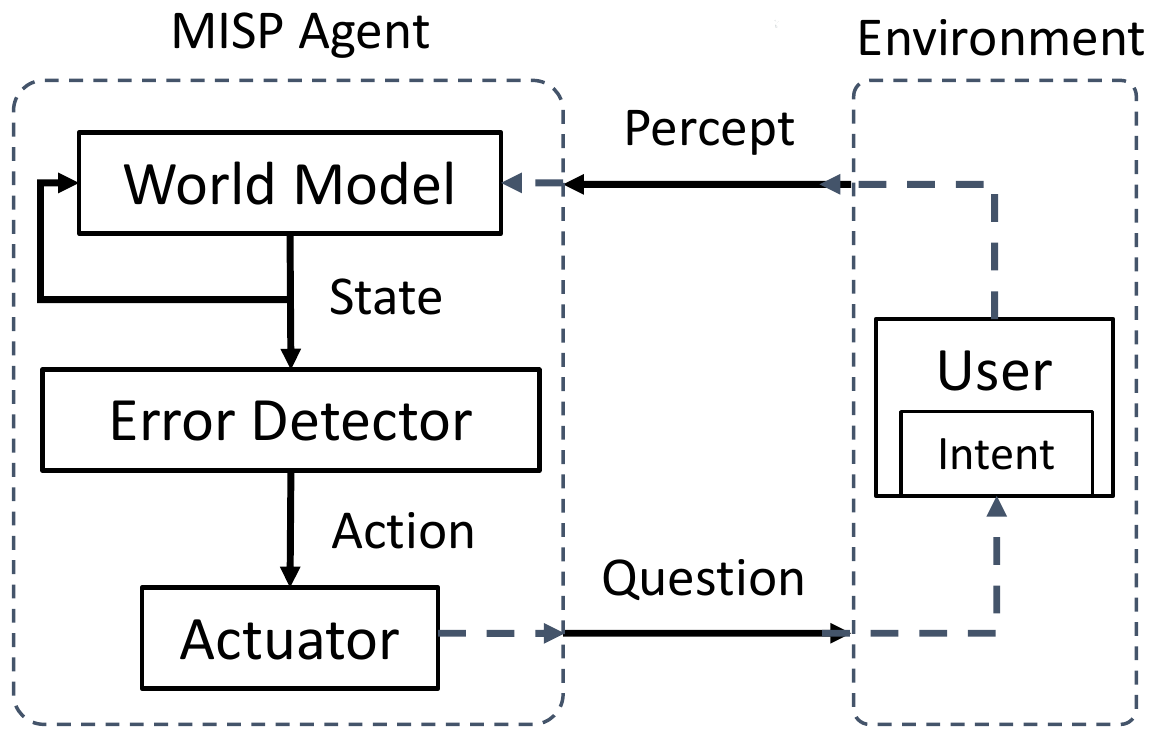}
    \caption{Model-based Interactive Semantic Parsing (MISP) framework.}
    \label{fig:misp}
\end{figure}

In response to these challenges, \emph{interactive semantic parsing} has been proposed recently as a practical solution, which includes human users in the loop to resolve utterance ambiguity, boost system accuracy, and improve user confidence via human-machine collaboration~\cite{li2014constructing, he2016human, chaurasia2017dialog, su2018natural, gur2018dialsql, yao2018interactive}.
For example, \citet{gur2018dialsql} built the DialSQL system to detect errors in a generated SQL query and request user selection on alternative options via dialogues.  Similarly, \citet{chaurasia2017dialog} and \citet{yao2018interactive} enabled semantic parsers to ask users clarification questions while generating an If-Then program.
\citet{su2018natural} showed that users overwhelmingly preferred an interactive system over the non-interactive counterpart for natural language interfaces to web APIs.
While these recent studies successfully demonstrated the value of interactive semantic parsing in practice, they are often bound to a certain type of formal language or dataset, and the designs are thus ad-hoc and not easily generalizable.  For example, DialSQL only applies to SQL queries on the WikiSQL dataset \cite{zhong2017seq2sql}, and it is non-trivial to extend it to other formal languages (e.g., $\lambda$-calculus) or even just to more complex SQL queries beyond the templates used to construct the dataset.

Aiming to develop a general principle for building interactive semantic parsing systems, in this work we propose \underline{m}odel-based \underline{i}nteractive \underline{s}emantic \underline{p}arsing (MISP), where the goal is to design a \emph{model-based intelligent agent} \cite{russell2009artificial} that can interact with users to complete a semantic parsing task.
\nop{
In this work, we propose a general formulation for interactive semantic parsing, the \underline{m}odel-based \underline{i}nteractive \underline{s}emantic \underline{p}arsing (MISP), where the goal is to design a \emph{model-based intelligent agent} \cite{russell2009artificial} that can interact with users to complete a semantic parsing task.
}
Taking an utterance (e.g., a natural language question) as input, the agent forms the semantic parse (e.g., a SQL query) in steps, potentially soliciting user feedback in some steps to correct parsing errors. 
As illustrated in Figure~\ref{fig:misp}, a MISP agent maintains its \textit{state} as the current semantic parse and, via an \textit{error detector}, decides whether and where human intervention is needed (the \textit{action}). This action is performed by a \textit{question generator} (the \textit{actuator}), which generates and presents to the user a human-understandable question. A core component of the agent is a \textit{world model} \cite{ha2018world} (hence \textit{model-based}), which incorporates user feedback from the environment and transitions to a new agent state (e.g., an updated semantic parse). This process repeats until a terminal state is reached.
Such a design endows a MISP agent with three crucial properties of\nop{to realize} interactive semantic parsing: (1) being \emph{introspective} of the reasoning process and knowing when it may need human supervision, (2) being able to \emph{solicit user feedback} in a human-friendly way, and (3) being able to \emph{incorporate user feedback} (through state transitions controlled by the world model).

The MISP framework provides several advantages for designing an interactive semantic parser compared to the existing ad-hoc studies. For instance, the whole problem is conceptually reduced to building three key components (i.e., the world model, the error detector, and the actuator), and can be handled and improved separately.  While each component may need to be tailored to the specific task, the general framework remains unchanged. 
In addition, the formulation of a model-based intelligent agent can facilitate the application of other machine learning techniques like reinforcement learning.

\nop{
In addition, the notion of state/action in a MISP agent naturally corresponds to the design of many modern semantic parsers \cite{dong2016language, xu2017sqlnet}, where partial semantic parses are extended in steps until the final output is constructed.  
}

\nop{
%
On the other hand, MIPS provides a unified formulation for existing (and potentially future) interactive semantic parsers. For example, DialSQL \cite{gur2018dialsql} can be viewed as a MISP agent operating over the space of SQL queries (state), with several neural networks composing the error detector and the world model. Similarly, the interactive parser in \cite{yao2018interactive} is a MISP agent that jointly learns the error detector and the state transition via reinforcement learning.
}

To better demonstrate the advantages of the MISP framework, we propose a simple yet remarkably effective instantiation for the text-to-SQL task. We show the effectiveness of the framework based on three base semantic parsers (SQLNet, SQLova and SyntaxSQLNet) and two datasets (WikiSQL and Spider). We empirically verified that with a small amount of targeted, test-time user feedback, interactive semantic parsers improve the accuracy by 10\% to 15\% absolute.  Compared to an existing interactive semantic parsing system, DialSQL \cite{gur2018dialsql}, our approach, despite its much simpler yet more general system design, achieves \nop{the same or }better parsing accuracy by asking only half as many questions.

\nop{

As will be demonstrated later, these makes MISP easily applicable to different tasks and different base semantic parsers.
As a case study to illustrate the framework, the second contribution of this work is introducing a \emph{simple yet remarkably effective} instantiation of the MISP agent in the setting of text-to-SQL, named ``MISP-SQL''. The major difference between MISP-SQL and DialSQL lies in the choice of the world model: In DialSQL, it is represented with a neural network trained additionally to the base parser (which is treated as a black box in DialSQL), leading to overheads on simulation data collection and network training. MISP-SQL, instead, uses the base semantic parser to model the world. Besides, MISP-SQL exploits the signals from the base parser (e.g., the parse probability) and adopts a straightforward threshold-based approach for error detection. Through experiments on two text-to-SQL datasets \cite{zhong2017seq2sql, yu2018spider}, we demonstrate the effectiveness of MISP-SQL when compared with the existing DialSQL system. 

}
\section{Background \& Related Work}
\label{sec:related_work}

\nop{
SY:  Just planning; most talking points probably are included already
 * semantic parsing --> should emphasize the state/action/RL kind of models & why it's naturally suitable to MISP
 * interactive semantic parsing --> relative new development, promising in practice, but the solution is quite ad-hoc (give concrete examples)
 * general dialog system --> how to differentiate it from ISP?
}

\begin{remark}[Semantic Parsing.]
Mapping natural language utterances to their formal semantic representations, semantic parsing has a wide range of applications, including question answering~\cite{berant2013semantic, dong2016language, finegan2018improving}, 
robot navigation~\cite{artzi2013weakly, thomason2015learning} and Web API calling~\cite{quirk2015language, su2018natural}.
The target application in this work is text-to-SQL, which has been popularized by the WikiSQL dataset~\cite{zhong2017seq2sql}.
One of the top-performing models on WikiSQL is SQLNet~\cite{xu2017sqlnet}, which leverages the pre-defined SQL grammar sketches on WikiSQL and solves the SQL generation problem via ``slot filling."  By augmenting SQLNet with a table-aware BERT encoder~\cite{devlin2018bert} and by revising the value prediction in \texttt{WHERE} clauses, SQLova~\cite{hwang2019comprehensive} advances further the state of the art.
Contrast to WikiSQL, the recently released Spider dataset~\cite{yu2018spider} focuses on complex SQL queries containing multiple keywords (e.g., \texttt{GROUP BY}) and may join multiple tables. To handle such complexity, \citet{yu2018syntaxsqlnet} proposed SyntaxSQLNet, a syntax tree network with modular decoders, which generates a SQL query by recursively calling a module following the SQL syntax. However, because of the more realistic and challenging setting in Spider, it only achieves 20\% in accuracy.\nop{One week before our submission, \citet{bogin2019representing} published a new result that improves the accuracy on Spider to 39.7\% using graph neural networks. In principle, their model can be used as the base semantic parser in MISP as well.}

{We experiment our MISP framework with the aforementioned three semantic parsers on both WikiSQL and Spider. The design of MISP allows naturally integrating them as the base parser. For example, when SQLNet fills a sequence of slots to produce a SQL query, a ``state'' in MISP corresponds to a partially generated SQL query and it transitions as SQLNet fills the next slot.}
\end{remark}

\begin{remark}[Interactive Semantic Parsing.]

To enhance parsing accuracy and user confidence in practical applications, interactive semantic parsing has emerged as a promising solution~\cite{li2014constructing, he2016human, chaurasia2017dialog, su2018natural, gur2018dialsql, yao2018interactive}.
Despite their effectiveness, existing solutions are somewhat ad-hoc and bound to a specific formal language and dataset. For example, DialSQL \cite{gur2018dialsql} is curated for WikiSQL, where SQL queries all follow the same and given grammar sketch. 
Similarly, \cite{yao2018interactive} relies on a pre-defined two-level hierarchy among components in an If-Then program and cannot generalize to formal languages with a deeper structure.
%
In contrast, MISP aims for a general design principle by explicitly identifying and decoupling important components, such as error detector, question generator and world model.  It also attempts to integrate and leverage a strong base semantic parser, and transforms it to a natural interactive semantic parsing system, which substantially reduces the engineering cost.   
\end{remark}

\section{Model-based Interactive Semantic Parsing} 
\label{sec:framework}

We now discuss the MISP framework (Figure~\ref{fig:misp}) in more detail.
Specifically, we highlight the function of each major building block and the relationships among them,
and leave the description of a concrete embodiment to Section~\ref{sec:MISP-SQL}.  

\begin{remark}[Environment.]
The \textit{environment} consists of a user with a certain intent, which corresponds to a semantic parse 
that the user expects the agent to produce. 
Based on this intent, the user gives an initial natural language utterance $u_0$ to start a semantic parsing \textit{session} and responds to any clarification question from the agent with feedback $u_t$ at interaction turn $t$.
\end{remark}

\begin{remark}[Agent State.]
The \textit{agent state} $s$ is an agent's internal interpretation of the environment based on all the available information. A straightforward design of the agent state is as the currently predicted semantic parse. It can also be endowed with meta information of the parsing process such as prediction probability or uncertainty to facilitate error detection.
\end{remark}

\begin{remark}[World Model.]
A key component of a MISP agent is its \textit{world model} \cite{ha2018world}, which compresses the historical percepts throughout the interaction and predicts the future based on the agent's knowledge of the world. 
More specifically, it models the \nop{transitioning}{transition} of the agent state, $p(s_{t+1} | s_t, u_t)$, where $u_t$ is the user feedback at step $t$ and $s_{t+1}$ is the new state. The \nop{transitioning}{transition} can be deterministic or stochastic.
\end{remark}


\begin{remark}[Error Detector.]
A MISP agent introspects its state and decides whether and where human intervention is needed. The \textit{error detector} serves this role.
Given the current state $s_t$ (optionally the entire interaction history) and a set of \textit{terminal states}, it decides on an \textit{action} $a_t$: If the agent is at a terminal state, it terminates the session, executes the semantic parse, and returns the execution results to the user; otherwise, it determines a span in the current semantic parse that is likely erroneous and passes it, along with necessary context information needed to make sense of the error span, to the actuator.  
\end{remark}

\begin{remark}[Actuator.]
An \textit{actuator} has a user-facing interface and realizes an agent's actions in a user-friendly way.
In practice, it can be a natural language generator (NLG)~\cite{he2016human, gur2018dialsql, yao2018interactive} or an intuitive graphical user interface~\cite{su2018natural,berant2019explaining}, or the two combined.
\end{remark}


\nop{In this section, we elaborate the MISP framework in Figure~\ref{fig:misp}. The core of this framework is an intelligent agent consisting of an error detector, an actuator as well as a world model. The agent maintains its own state as the current (partial) semantic parse and updates the state while interacting with the environment. We will first introduce the environment then formally define each agent element.}

\nop{
It is the agent's internal interpretation of the environment (i.e., user intent). In DialSQL \cite{gur2018dialsql}, it is the SQL query after being validated by the user for $t$ times (i.e., after $t$ times of interactions). In our instantiated text-to-SQL system MISP-SQL (to be introduced in Section~\ref{sec:MISP-SQL}), it is the partial SQL query after $t$ steps of user-agent interactions.
}

\nop{
In practice, this world model can be instantiated with any machine learning model. For example, in DialSQL, the world model is a neural network learned from scratch on a collected simulation data, while in \cite{chaurasia2017dialog, yao2018interactive}, it is the same base semantic parser with or without fine-tuning. 
}

\section{MISP-SQL: An Instantiation of MISP for Text-to-SQL}
\label{sec:MISP-SQL}



\nop{
  The opening of the section 4 should cover the following:
  * The target application of a MISP agent in this work (text-to-SQL)
  * The use of a base semantic parser in the embodiment (add a footnote on other ways of implementing a MISP agent, especially when a base semantic parser is not available or given?)
  * Explaining Figure 2
}

\nop{
To illustrate the MISP framework, in this section, we instantiate it under the setting of text-to-SQL and present an interactive SQL parsing system called ``MISP-SQL'' (Figure~\ref{fig:MISP-SQL}). 
}


Under the MISP framework, we design an interactive semantic parsing system (Figure~\ref{fig:MISP-SQL}), named MISP-SQL, for the task of text-to-SQL translation. MISP-SQL assumes a base text-to-SQL parser and leverages it to design the world model and the error detector.\nop{While leveraging an existing semantic parser is perhaps the most efficient way in terms of engineering cost, the MISP framework can still be adopted when such semantic parsers do not exist.\nop{ See Section~\ref{sec:future} for more discussion.}} The world model is essentially a wrapper that takes the user input and changes the behavior of the base semantic parser (e.g., by changing the probability distribution or removing certain prediction paths). The error detector makes decisions based on the \textit{uncertainty} of the predictions: if the parser is uncertain about a prediction, it is more likely to be an error. The actuator is a template-based natural language question generator developed for the general SQL language. Figure~\ref{fig:MISP-SQL} shows an example of the MISP-SQL agent.


\subsection{Agent State}

For ease of discussion, we assume the base parser generates the SQL query by predicting a sequence of SQL components,\footnote{In practice this assumption may not be necessary as long as there is a reasonable way to chunk the semantic parse to calculate uncertainty and formulate clarification questions.}
as in many state-of-the-art systems~\cite{xu2017sqlnet, wang2018pointing, yu2018typesql, hwang2019comprehensive}.
Agent state $s_t$ is thus defined as a \textit{partial} SQL query, i.e., $s_t$=\{$o_1, o_2, ..., o_t$\}, where $o_t$ is the predicted SQL component at time step $t$, such as \texttt{SELECT place} in Figure~\ref{fig:MISP-SQL}.  What constitutes a SQL component is often defined differently in different semantic parsers, but typically dictated by the SQL syntax. 
To support introspection and error detection, each prediction is associated with its uncertainty, which is discussed next.

\nop{
Following the custom of traditional one-shot text-to-SQL parsers \cite{xu2017sqlnet, wang2018pointing, yu2018typesql, hwang2019comprehensive}, we assume that a base SQL parser generates a SQL query via a sequence of predictions on SQL components. An agent state $s_t$ is thus defined as a \textit{partially} decoded SQL query, i.e., $s_t$=\{$o_1, o_2, ..., o_t$\}, where $o_t$ is the predicted SQL component at time step $t$.
}

\subsection{Error Detector}\label{subsec:sql_ed}

The error detector in MISP-SQL is introspective and greedy. It is introspective because it examines the uncertainty of the predictions as opposed to the predictions themselves. On the other hand, it is greedy because its decisions are solely based on the last prediction $o_t$ instead of the entire state $s_t$.

We experiment with two uncertainty measures, based on the probability of $o_t$ estimated by the base semantic parser, as well as its standard deviation under Bayesian dropout~\cite{gal2016dropout}, respectively.


\begin{remark}[Probability-based Uncertainty.]
Intuitively if the base semantic parser gives a low probability to the top prediction at a step, it is likely uncertain about the prediction. Specifically, we say a prediction $o_t$ needs user clarification if its probability is lower than a threshold $p^*$, i.e.,
$$p(o_t) < p^*.$$
This strategy is shown to be strong in detecting misclassified and out-of-distribution examples \cite{hendrycks17baseline}.
\end{remark}

\begin{figure}
    \centering
    \includegraphics[width=\columnwidth]{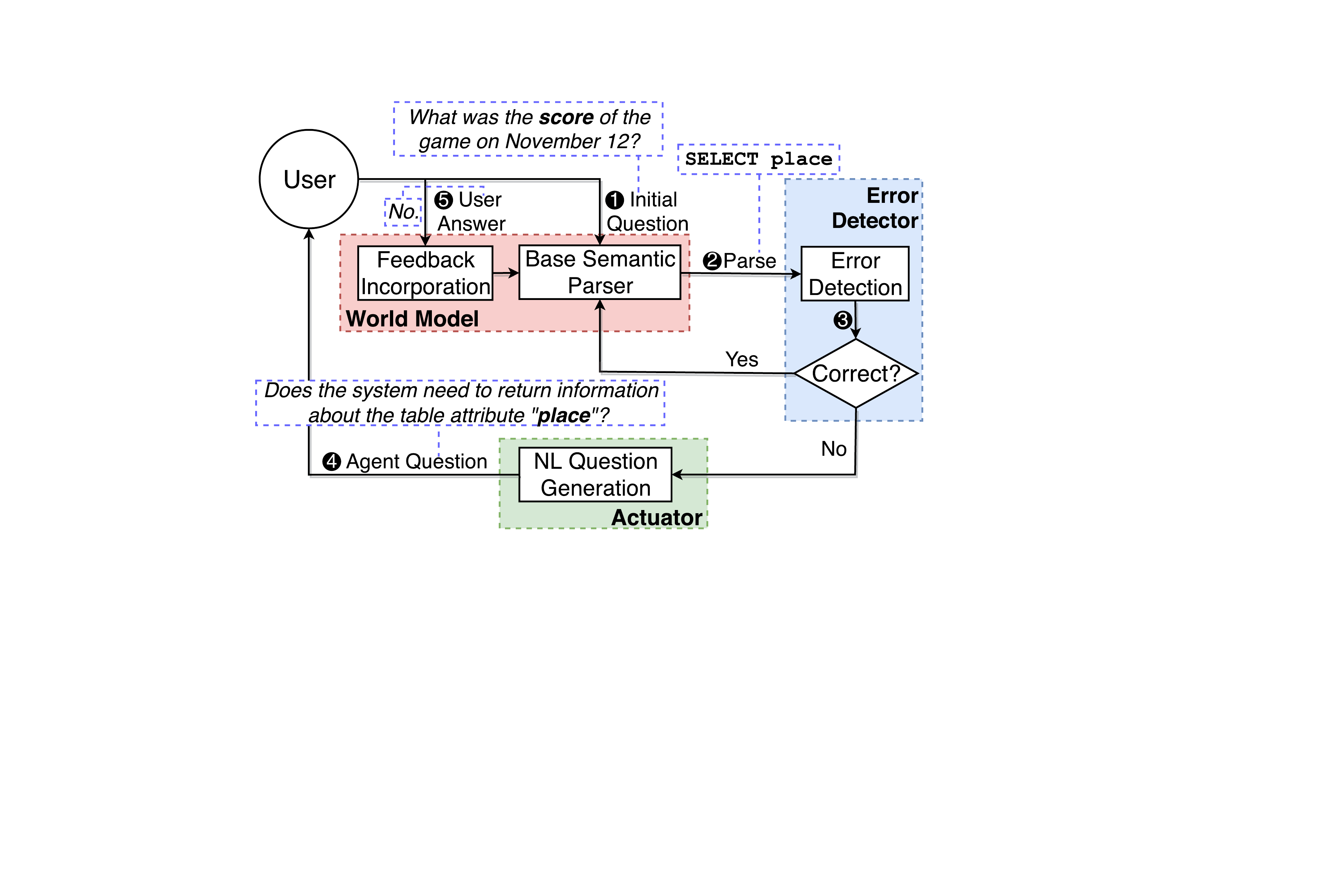}
    \caption{MISP-SQL Agent. The base semantic parser incrementally parses the user question (Step 1) into a SQL query by first selecting a column from the table (Step 2). This partial parse is examined by the error detector (Step 3), who determines that the prediction is incorrect (because the uncertainty is high) and triggers the actuator to ask the user a clarification question (Step 4). The user feedback is then incorporated into the world model (Step 5) to update the agent state. If the prediction was correct, Step 2 would be repeated to continue the parsing.}
    \label{fig:MISP-SQL}
    \vspace{-10pt}
\end{figure}

\begin{remark}[Dropout-based Uncertainty.]
Dropout \cite{srivastava2014dropout} has been used as a Bayesian approximation for estimating model uncertainty \cite{gal2016dropout} in several tasks \cite{dong2018confidence, siddhant2018deep, xiao2018quantifying}. Different from its standard application to prevent models from overfitting in training time, we use it at test time to measure model uncertainty, similar to \cite{dong2018confidence}. The intuition is that if the probability on a prediction varies dramatically (as measured by the standard deviation) across different perturbations under dropout, the model is likely uncertain about it.
Specifically, the uncertainty on prediction $o_t$ is calculated as:
$$\textrm{STDDEV} \{p(o_t|W_i)\}^N_{i=1}, $$
where $W_i$ is the parameters of the base semantic parser under the $i$-th dropout perturbation, and the uncertainty score is the standard deviation of the prediction probabilities over $N$ random passes. We say $o_t$ needs user clarification if its uncertainty score is greater than a threshold $s^*$.
\end{remark}

\begin{remark}[Terminal State.]
The only terminal state is when the base semantic parser indicates end of parsing.
\end{remark}

\nop{As shown in Figure~\ref{fig:MISP-SQL}, when a prediction is considered likely wrong, the agent passes it (as well as its context) to the actuator for requesting user validation; otherwise, the base parser continues the SQL generation.}

\begin{table*}[t]
    \centering\small
    \begin{tabular}{c}\toprule
    \textbf{[Lexicon]}\\[1mm]
    is greater than\textbar equals to\textbar is less than $\rightarrow$ \textsc{Op}[\texttt{$>$}\textbar \texttt{$=$}\textbar \texttt{$<$}]\\
    sum of values in\textbar average value in\textbar number of\textbar minimum value in\textbar maximum value in $\rightarrow$ \textsc{Agg}[\texttt{sum}\textbar\texttt{avg}\textbar\texttt{count}\textbar\texttt{min}\textbar\texttt{max}] \\
    \midrule
    \textbf{[Grammar]}\\[1mm]
    ``$col$'' $\rightarrow$ \textsc{Col}[$col$]\\
    Does the system need to return information about \textsc{Col}[$col$] ? $\rightarrow$ Q[$col\|$\texttt{SELECT} $agg?$ $col$]\\
    Does the system need to return \textsc{Agg}[$agg$] \textsc{Col}[$col$] ? $\rightarrow$ Q[$agg\|$\texttt{SELECT} $agg$ $col$] \\
    Does the system need to return a value \underline{after} any mathematical calculations on \textsc{Col}[$col$] ? $\rightarrow$ Q[$agg$=None$\|$\texttt{SELECT} $col$] \\
    Does the system need to consider any conditions about \textsc{Col}[$col$] ? $\rightarrow$ Q[$col\|$\texttt{WHERE} $col$ $op$ $val$] \\
    The system considers the following condition: \textsc{Col}[$col$] \textsc{Op}[$op$] a value. Is this condition correct? $\rightarrow$ Q[$op\|$\texttt{WHERE} $col$ $op$ $val$]\\
    The system considers the following condition: \textsc{Col}[$col$] \textsc{Op}[$op$] $val$. Is this condition correct? $\rightarrow$ Q[$val\|$\texttt{WHERE} $col$ $op$ $val$]\\
    \bottomrule
    \end{tabular}
    \caption{Domain-general lexicon and grammar for NL generation in MISP-SQL (illustrated for WikiSQL; a more comprehensive grammar for Spider can be found in Appendix~\ref{app:ext_spider}).}
    \label{tab:grammar}
    \vspace{-10pt}
\end{table*}

\subsection{Actuator: An NL Generator}\label{subsec:NLG}

The MISP-SQL agent performs its action (e.g., validating the column ``\textit{place}'') via asking users \textit{binary} questions, hence the actuator is a natural language generator (NLG). Although there has been work on describing a SQL query with an NL statement \cite{koutrika2010explaining, ngonga2013sorry, iyer2016summarizing, xu2018sql}, {few work studies generating \textit{questions} about a certain SQL component in a systematic\nop{principled} way.}
\nop{to the best of our knowledge, we are the first to generate \textit{questions} on a certain SQL component \fromH{So DialSQL does not generate questions or not on a component?}.}

Inspired by \cite{koutrika2010explaining, wang2015building}, we define a rule-based NLG, which consists of a seed lexicon and a grammar for deriving questions. Table~\ref{tab:grammar} shows rules covering SQL queries on WikiSQL \cite{zhong2017seq2sql}. 
The seed lexicon defines NL descriptions for basic SQL elements in the form of ``$n \rightarrow t[p]$'', where $n$ is an NL phrase, $t$ is a pre-defined syntactic category and $p$ is either an aggregator (e.g., \texttt{avg}) or an operator (e.g., $>$). For example, ``is greater than $\rightarrow$ \textsc{Op}[$>$]'' specifies a phrase ``is greater than'' to describe the operator ``$>$''.
In MISP-SQL, we consider four syntactic categories: \textsc{Agg} for aggregators, \textsc{Op} for operators, \textsc{Col} for columns and \textsc{Q} for generated questions. However, it can be extended with more lexicon entries and grammar rules to accommodate more complex SQL in Spider \cite{yu2018spider}, which we show in Appendix~\ref{app:ext_spider}.

\begin{figure}[t]
    \centering
    \includegraphics[width=\linewidth]{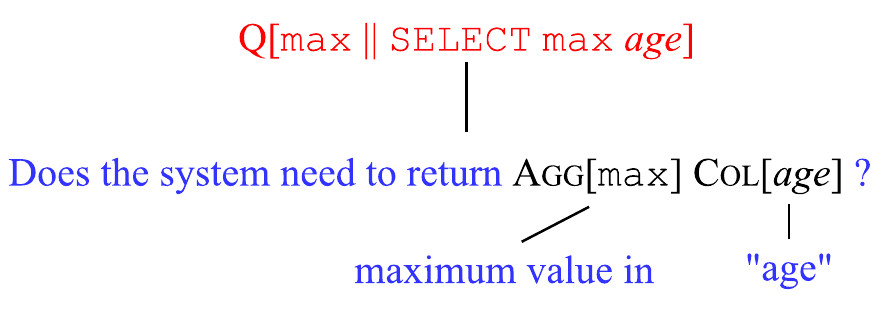}
    \caption{Deriving an NL question about the aggregator \texttt{max} in the clause ``\texttt{SELECT} \texttt{max}(age)'' from the rooted Q-typed item.}
    \label{fig:derivation}
    \vspace{-2mm}
\end{figure}

The grammar defines rules to derive questions. Each column is described by itself (i.e., the column name). Rules associated with each Q-typed item ``Q[$v\|$\texttt{Clause}]'' constructs an NL question asking about $v$ in \texttt{Clause}. The \texttt{Clause} is the necessary context to formulate meaningful questions. Figure~\ref{fig:derivation} shows a derivation example. Note that, both the lexicon and the grammar in our system are domain-agnostic in the sense that it is not specific to any database. {Therefore, it can be reused for new domains in the future.} Database-specific rules, such as naming each column with a more canonical phrase (rather than the column name), are also possible.

\subsection{World Model}\label{subsec:sql_model}
The agent incorporates user feedback and updates its state with a world model. Different from DialSQL which trains an additional neural network, the MISP-SQL agent directly employs the base semantic parser to transition states, which saves additional training efforts.

As introduced in Section~\ref{subsec:NLG}, the agent raises a binary question to the user about a predicted SQL component $o_t$. Therefore, the received user feedback either confirms the prediction or negates it. In the former case, the state is updated by proceeding to the next decoding step, i.e., $s_{t+1}$=\{$o_1$, ..., $o_t$, $o_{t+1}$\}, where $o_{t+1}$ is\nop{a newly predicted} {the predicted next} component {and $s_{t+1}$ shows the updated partial parse}. In the latter case, the user feedback is incorporated to constrain the search space of the base parser (i.e., forbidding the parser from making the same wrong prediction), based on which the parser refreshes its prediction and forms a new state $s_{t+1}$=\{$o_1$, ..., $o_{t-1}$, $o_{t+1}$\}, where $o_{t+1}$ is \nop{the updated prediction}{a predicted alternative} to replace $o_t$\nop{\fromH{i feel the the time step here is confusing. if the user continues to negate $o_{t+1}$, will the state become $s_{t+1}$=\{...$o_{t+2}$\}? basically the $o_{t+1}$ here means a different thing from that in the previous sentence.}\fromZ{[Ziyu] yes; yes they correspond to different components, but the time step is right. I revised the previous sentences a bit}}. To avoid being trapped in a large search space, for each SQL component, we consider a maximum number of $K$ \nop{questions} {alternatives (in addition to the original prediction)} {to solicit user feedback on.}

\section{Experiments} \label{sec:exp}

We apply our approach to the task of mapping natural language questions to SQL queries.  
In this section, we first describe the basic setup, including the datasets and the base semantic parsers,
followed by the system results on both simulated and real users.

\begin{table*}[t!]
    \centering\small
    \begin{tabular}{l c c c c c c}\toprule
         & \multicolumn{3}{c}{\textbf{SQLNet}} & \multicolumn{3}{c}{\textbf{SQLova}}\\
        \textbf{System} & \textbf{Acc\textsubscript{qm}} & \textbf{Acc\textsubscript{ex}} & \textbf{Avg. \#q} &  \textbf{Acc\textsubscript{qm}} & \textbf{Acc\textsubscript{ex}} & \textbf{Avg. \#q} \\\midrule
        no interaction & 0.615 & 0.681 & N/A & 0.797 & 0.853 & N/A  \\\midrule
        DialSQL & 0.690 & N/A & 2.4\nop{4.77} & N/A & N/A & N/A  \\\midrule
        MISP-SQL\textsuperscript{Unlimit10} & 0.932 & 0.948 & 7.445 & 0.985 & 0.991 & 6.591 \\
        MISP-SQL\textsuperscript{Unlimit3} & 0.870 & 0.900 & 7.052 & 0.955 & 0.974 & 6.515 \\\midrule
        MISP-SQL\textsuperscript{$p^*$=0.95} & 0.782 & 0.824 & 1.713 & 0.912 & 0.939 & 0.773 \\
        MISP-SQL\textsuperscript{$p^*$=0.8} & 0.729 & 0.779 & 1.104 & 0.880 & 0.914 & 0.488\\
        \bottomrule
    \end{tabular}
    \caption{Simulation evaluation of MISP-SQL (based on SQLNet or SQLova) on WikiSQL Test set. ``MISP-SQL\textsuperscript{$p^*$=X}'' denotes our agent with probability-based error detection (threshold at X). ``MISP-SQL\textsuperscript{UnlimitK}'' denotes a variant that asks questions for every component, with up to {$K+1$ questions per component}.}
    \label{tab:sim_results}
    \vspace{-10pt}
\end{table*}

\subsection{Experimental Setup}
We evaluate our proposed MISP-SQL agent on WikiSQL~\cite{zhong2017seq2sql}, which contains 80,654 hand-annotated pairs of $\langle$NL question, SQL query$\rangle$, distributed across 24,241 tables from Wikipedia. Our experiments follow the same data split as in~\cite{zhong2017seq2sql}.

We experiment MISP-SQL with two base semantic parsers: SQLNet \cite{xu2017sqlnet} and SQLova \cite{hwang2019comprehensive}\nop{, where the latter is currently the best open-sourced model on WikiSQL}. Unlike in DialSQL's evaluation \cite{gur2018dialsql}, we do not choose Seq2SQL \cite{zhong2017seq2sql} as a base parser but SQLova instead, because it achieves similar \nop{mediocre} performance as SQLNet while SQLova is currently the best open-sourced model on WikiSQL, which can give us a more comprehensive evaluation. 
For each of the two base semantic parsers, we test our agent with two kinds of error detectors, based on prediction probability and \nop{model uncertainty}{Bayesian dropout}, respectively (Section~\ref{subsec:sql_ed}).  We tune the threshold $p^*$ within 0.5 $\sim$ 0.95 and $s^*$ within 0.01 $\sim$ 0.2. Particularly for uncertainty-based detection measured by Bayesian dropout, the number of passes $N$ is set to 10, with a dropout rate 0.1. The dropout layers are applied at the same positions as when each semantic parser is trained. When the agent interacts with users, the maximum number of \nop{questions}{alternative options (in addition to the original prediction)} per component, $K$, is set to 3. {If the user negates all the $K+1$ predicted candidates, the agent will keep the original prediction, as in \cite{gur2018dialsql}.}



\subsection{Simulation Evaluation}
In simulation evaluation, each agent interacts with a \textit{simulated} user, who \nop{. The user }gives a yes/no answer based on the ground-truth SQL query. 
\nop{
The simulated user can also stop the interaction process when the agent consecutively asks about three wrong predictions without correctly fixing them.
}
\nop{If the agent fails to correct its predictions on three consecutively asked wrong SQL components}{If the agent fails to correct its predictions in three consecutive interaction turns}, the user will leave the interaction early and the agent has to finish the remaining generation without further help from the user.

\paragraph{Overall Comparison.}
\nop{We first compare MISP-SQL with baselines, either the two base semantic parsers without interactions or the existing DialSQL system, in Table~\ref{tab:sim_results}.} 
{We first compare MISP-SQL with the two base semantic parsers without interactions in Table~\ref{tab:sim_results}. For SQLNet, we also compare our system with the reported performance of DialSQL \cite[Table 4]{gur2018dialsql}. However, since DialSQL is not open-sourced and it is not easy to reproduce it, we are unable to adapt it to SQLova for more comparisons.} Following \cite{xu2017sqlnet, hwang2019comprehensive}, we evaluate the SQL query match accuracy (``\textbf{Acc\textsubscript{qm}}'', after converting the query into its canonical form) and the execution accuracy (``\textbf{Acc\textsubscript{ex}}'') of each agent. ``\textbf{Avg. \#q}'' denotes the average number of questions per query. For any base parser, MISP-SQL improves their performance by interacting with users. Particularly for SQLNet, MISP-SQL outperforms the DialSQL system with only \emph{half} the number of questions (1.104 vs. 2.4), 
and has a much simpler design without the need of training an extra model (besides training the base parser, which DialSQL needs to do as well). Our agent can even boost the strong performance of SQLova from 85\% to 94\% in execution accuracy, with merely 0.773 questions per query.\nop{\footnote{It is not trivial to re-implement DialSQL and its source code is not released. Hence, we did not adapt it to SQLova.}}

We also present an ``upper-bounded'' accuracy of our agent, when it does not adopt any error detector and asks questions about \textit{every} \nop{prediction }{component} with at most 10 (``MISP-SQL\textsuperscript{Unlimit10}'') or 3 (``MISP-SQL\textsuperscript{Unlimit3}'') alternatives. 
Interestingly, even for the weaker SQLNet parser, most true predictions have already been contained within the top 10 options (giving 0.932 query match accuracy). When equipped with the stronger SQLova parser, the agent has a potential to boost the execution accuracy to around 100\% by considering only the top 3 options of every prediction. 
The complete results can be found in Appendix~\ref{app:sim_results}.

\begin{figure}[t!]
    \centering
    \includegraphics[width=\linewidth]{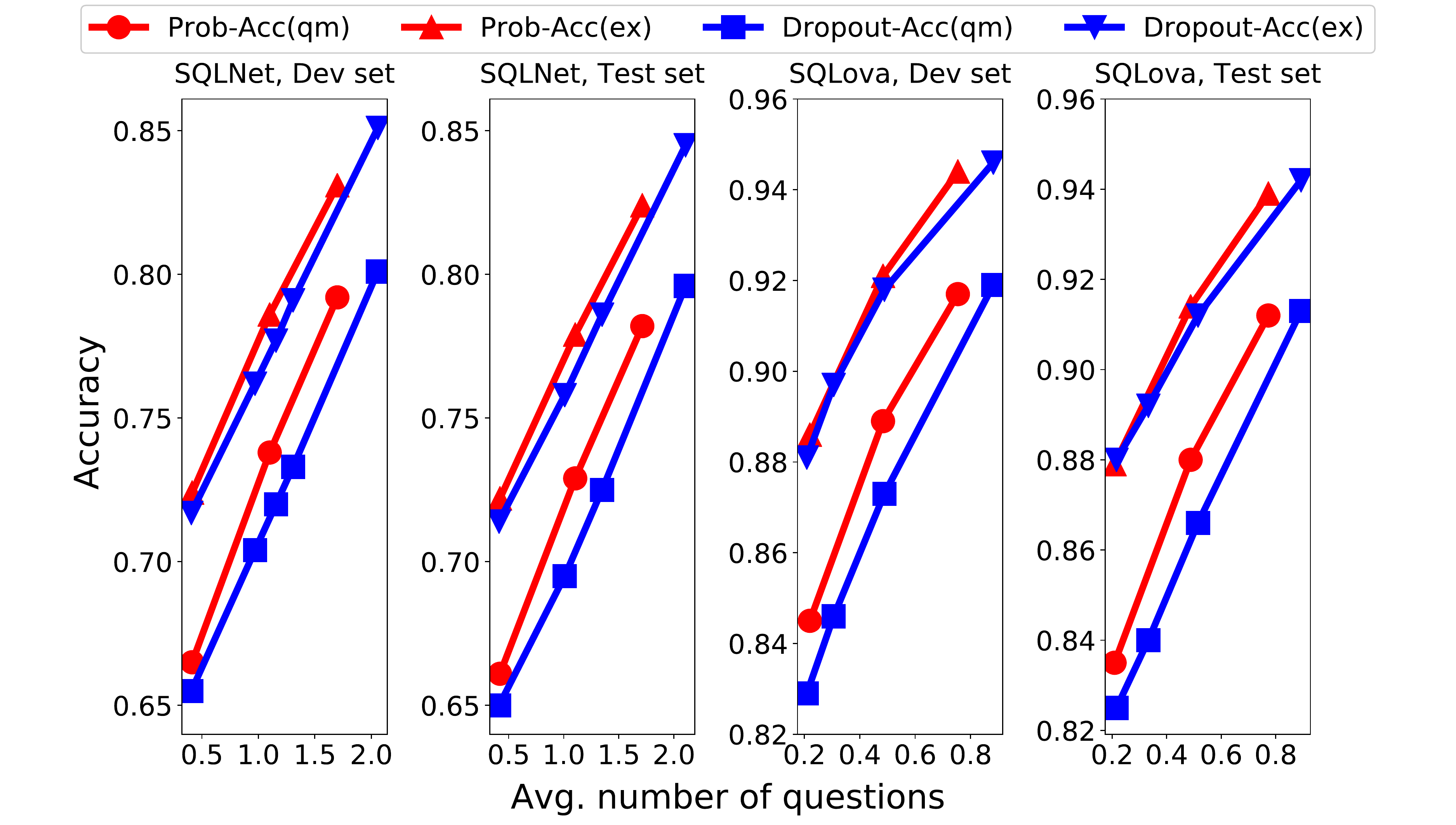}
    \caption{Comparison of probability- and dropout-based error detection.}
    \label{fig:sim_results}
    \vspace{-5pt}
\end{figure}

\paragraph{Error Detector Comparison.}
We then compare the probability-based and dropout-based error detectors in Figure~\ref{fig:sim_results}, where each marker indicates the agent's accuracy and the average number of questions it needs under a certain error detection threshold. Consistently for both SQLNet and SQLova, the probability-based error detector can achieve the same accuracy with a lower number of questions than the dropout-based detector. It is also observed that this difference is greater in terms of query match accuracy, around 0.15 $\sim$ 0.25 for SQLNet and 0.1 $\sim$ 0.15 for SQLova. 
A more direct comparison of various settings under the same average number of questions can be found in Appendix~\ref{app:error_detector}.

To better understand how each kind of error detectors works, we investigate the portion of questions that each detector spends on \textit{right} predictions (denoted as ``\textbf{Q\textsubscript{r}}''). An ideal system should ask fewer questions on right predictions while identify more truly incorrect predictions to fix the mistakes.
We present the question distributions of the various systems in Table~\ref{tab:question_cat}. One important conclusion drawn from this table is that probability-based error detection is much more effective on identifying incorrect predictions. Consider the system using probability threshold 0.5 for error detection (i.e., ``$p^*$=0.5'') and the one using dropout-based error detector with a threshold 0.2 (i.e., ``$s^*$=0.2'') on SQLNet. When both systems ask around the same number of questions during the interaction, the former spends only 16.9\% of unnecessary questions on correct predictions (Q\textsubscript{r}), while the latter asks twice amount of them (32.1\%). Similar situations are also observed for SQLova. It is also notable that, when the probability threshold is lower (which results in a fewer total number of questions), the portion of questions on right actions drops significantly (e.g., from 23.0\% to 16.9\% when the threshold changes from 0.8 to 0.5 on SQLNet). However, this portion remains almost unchanged for dropout-based error detection.

\subsection{Extend to Complex SQL Generation}
A remarkable characteristic of MISP-SQL is its generalizability, as it makes the best use of the base semantic parser and requires no extra model training. To verify it, we further experiment MISP-SQL on the more complex text-to-SQL dataset ``Spider'' \cite{yu2018spider}. The dataset consists of 10,181 questions on multi-domain databases, where SQL queries can contain complex keywords such as \texttt{GROUP BY} or join several tables. We extend the NLG lexicon and grammar (Section~\ref{subsec:NLG}) to accommodate this complexity, with details shown in Appendix \ref{app:ext_spider}.

\begin{table}[t]
    \centering \small
    \scalebox{0.92}{
    \begin{tabular}{l c c l c c}
    \toprule
     \multicolumn{3}{c}{\textbf{SQLNet}} & \multicolumn{3}{c}{\textbf{SQLova}}\\
    \midrule
        \textbf{System} & \textbf{Avg. \#q} & \textbf{Q\textsubscript{r}\%} &  \textbf{System} & \textbf{Avg. \#q} & \textbf{Q\textsubscript{r}\%} \\
    \midrule
        $p^*$=0.8 & 1.099\nop{9,248} & 23.0\% & $p^*$=0.8 & 0.484\nop{4,078} & 28.9\% \\
        $p^*$=0.5 & 0.412\nop{3,463} & 16.9\% & $p^*$=0.5 & 0.220\nop{1,852} & 18.4\% \\
        $s^*$=0.07 & 1.156\nop{9,733} & 34.5\% & $s^*$=0.03 & 0.489\nop{4,118} & 50.4\% \\
        $s^*$=0.2 & 0.406\nop{3,416} & 32.1\% & $s^*$=0.05 & 0.306\nop{2,577} & 52.5\% \\
    \bottomrule
    \end{tabular}
    }
    \caption{Portion of interaction questions on right predictions (Q\textsubscript{r}\%) for each agent setting {on WikiSQL Dev set} (smaller is better). ``$p^*$/$s^*$=X'' denotes our agent with probability/dropout-based error detection (threshold at X).}
    \label{tab:question_cat}
    \vspace{-5pt}
\end{table}

We adopt \nop{the state-of-the-art }SyntaxSQLNet \cite{yu2018syntaxsqlnet} as the base parser.\footnote{{We chose SyntaxSQLNet because it was the best model by the paper submission time. In principle, our framework can also be applied to more sophisticated parsers such as \cite{bogin2019representing, guo-etal-2019-towards}.}} In our experiments, we follow the same database split as in \cite{yu2018spider} and report the Exact Matching accuracy (``\textbf{Acc\textsubscript{em}}'') on Dev set.\footnote{We do not report results on Spider test set since it is not publicly available.} Other experimental setups remain the same as when evaluating MISP-SQL on WikiSQL. Table~\ref{tab:syntaxSQL} shows the results.

We first observe that, via interactions with simulated users, MISP-SQL improves SyntaxSQLNet by 10\% accuracy with reasonably 3 questions per query. However, we also realize that, unlike on WikiSQL, in this setting, the probability-based error detector requires more questions than the Bayesian uncertainty-based detector. This can be explained by the inferior performance of the base SyntaxSQLNet parser (merely 20\% accuracy without interaction). In fact, the portion of questions that the probability-based detector spends on right predictions (Q\textsubscript{r}) is still half of that the dropout-based detector asks (12.8\% vs. 24.8\%). However, it wastes around 60\% of questions on unsolvable wrong predictions. This typically happens when the base parser is not strong enough, i.e., cannot rank the true option close to the top, or when there are unsolved wrong precedent predictions (e.g., in ``\texttt{WHERE} $col$ $op$ $val$'', when $col$ is wrong, whatever $op$/$val$ following it is wrong). This issue can be alleviated when more advanced base parsers are adopted in the future.

\begin{table}[t]
    \centering\small
    \begin{tabular}{lcc}\toprule
        \textbf{System} & \textbf{Acc\textsubscript{em}} & \textbf{Avg. \#q}\\\midrule
        no interaction & 0.190 &  N/A\\\midrule
        MISP-SQL\textsuperscript{Unlimit10} & 0.522 & 14.878 \\
        MISP-SQL\textsuperscript{Unlimit3} & 0.382 & 11.055 \\\midrule
        MISP-SQL\textsuperscript{$p^*$=0.95} & 0.300 & 3.908 \\
        MISP-SQL\textsuperscript{$p^*$=0.8} & 0.268 & 3.056 \\
        MISP-SQL\textsuperscript{$s^*$=0.01} & 0.315 & 3.815 \\
        MISP-SQL\textsuperscript{$s^*$=0.03} & 0.290 & 2.905 \\\bottomrule
    \end{tabular}
    \caption{Simulation evaluation of MISP-SQL (built on SyntaxSQLNet) on Spider Dev set.}
    \label{tab:syntaxSQL}
    \vspace{-5pt}
\end{table}

\subsection{Human Evaluation}
\label{sec:human_exp}

We further conduct human user study to evaluate the MISP-SQL agent. Our evaluation setting largely follows \citet{gur2018dialsql}. For each base semantic parser, we randomly sample 100 examples from the corresponding dataset (either WikiSQL Test set or Spider Dev set) and ask three human evaluators, who are graduate students with only rudimentary knowledge of SQL based on our survey, to work on each example and then report the averaged results. We present to the evaluators the initial natural language question and allow them to view the table headers to better understand the question intent. On Spider, we also show the name of the database tables. We select error detectors based on the simulation results: For SQLNet and SQLova, we equip the agent with a probability-based error detector (threshold at 0.95); for SyntaxSQLNet, we choose a Bayesian uncertainty-based error detector (threshold at 0.03). As in the simulation evaluation, we cannot directly compare with DialSQL in human evaluation because the code is not yet publicly available.

\nop{For each of the above three settings, we randomly sample 100 examples from either WikiSQL Test set or Spider Dev set. Following \cite{gur2018dialsql}, we let three users to work on the same example and report the average performance. For each example, we present users the table headers and the initial natural language question. On Spider dataset, we also show the names of tables in the database. Users are then instructed to answer questions prompted by the system. }

Table~\ref{tab:human_eval} shows the results.
In all settings, MISP-SQL improves the base parser's performance, demonstrating the benefit of involving human interaction. However, we also notice that the gain is not as large as in simulation, especially on SQLova. Through interviews with the human evaluators, we found that the major reason is that they sometimes had difficulties understanding the true intent of some test questions that are ambiguous, vague, or contain entities they are not familiar with. We believe this reflects a general challenge of setting up human evaluation for semantic parsing that is close to the real application setting, \nop{hence}{and thus set forth} the following discussion.

\begin{table}[t]
    \centering\small
    \begin{tabular}{lccc}
    \toprule
        \textbf{System} & \textbf{Acc\textsubscript{qm/em}} & \textbf{Acc\textsubscript{ex}} & \textbf{Avg. \#q}\\\midrule
        \multicolumn{4}{c}{\textbf{SQLNet}}\\\midrule
        no interaction & 0.580 & 0.660 & N/A \\
        MISP-SQL (simulation) & 0.770 & 0.810 & 1.800 \\
        MISP-SQL (real user) & 0.633 & 0.717 & {1.510} \\\midrule
        \multicolumn{4}{c}{\textbf{SQLova}}\\\midrule
        no interaction & 0.830 & 0.890 & N/A \\
        MISP-SQL (simulation) & 0.920 & 0.950 & 0.550 \\
        MISP-SQL (real user) & 0.837 & 0.880 & 0.533 \\
        \hspace{2mm} + w/ full info. & 0.907 & 0.937 & 0.547\\\midrule
        \multicolumn{4}{c}{\textbf{SyntaxSQLNet}}\\\midrule
        no interaction & 0.180 & N/A & N/A \\
        MISP-SQL (simulation) & 0.290 & N/A & 2.730 \\
        MISP-SQL (real user) & 0.230 & N/A & 2.647 \\
    \bottomrule
    \end{tabular}
    \caption{Human evaluation on 100 random examples for MISP-SQL agents based on SQLNet, SQLova and SyntaxSQLNet, respectively.}
    \label{tab:human_eval}
    \vspace{-5pt}
\end{table}

\subsection{Discussion on Future Human Evaluation}

Most human evaluation studies for (interactive) semantic parsers so far \cite{chaurasia2017dialog, gur2018dialsql,su2018natural, yao2018interactive} use pre-existing test questions (e.g., from datasets like WikiSQL). However, this introduces an undesired discrepancy, that is, human evaluators may not necessarily be able to understand the true intent of the given questions in an faithful way, especially when the question is ambiguous, vague, or containing unfamiliar entities. 

This discrepancy is clearly manifested in our human evaluation with SQLova (Table~\ref{tab:human_eval}). {When the base parser is strong}, many of the remaining incorrectly parsed questions are challenging not only for the base parser but also for human evaluators. We manually examined the situations where evaluators made a different choice than the simulator and found that 80\% of such choices happened when the initial question is ambiguous or the gold SQL annotation is wrong. For example, for the question ``\textit{name the city for kanjiža}'' it is unlikely for human evaluators to know that ``kanjiža'' is an ``Urban Settlement'' without looking at the table content or knowing the specific background knowledge beforehand. This issue has also been reported as the main limitation to further improve SQLova \cite{hwang2019comprehensive}, which could in principle be resolved by human interactions if the users \nop{know what they are asking}{have a clear and consistent intent in mind}. To verify this, we conduct an additional experiment with SQLova where human evaluators can view the table content as well as the gold SQL query before starting the interaction to better understand the true intent (denoted as ``w/ full info'' in Table~\ref{tab:human_eval}). As expected, the MISP-SQL agent performs much better (close to simulation) when users know what they are asking. It further confirms that a non-negligible part of the accuracy gap between simulation and human evaluation is due to human evaluators not fully understanding the question intent and giving false feedback.

To {alleviate}\nop{bridge} this discrepancy, a common practice is to show human evaluators the schema of the underlying database, as \citet{gur2018dialsql} and we did (Section~\ref{sec:human_exp}), but it is \nop{often }still insufficient, especially for entity-related issues (e.g., ``kanjiža''). On the other hand, while exposing human evaluators to table content helps resolve the entity-related issues, \nop{overexposure also introduces}{it is likely to introduce} undesired biases in favor of the system under test {(i.e., ``overexposure'')}, since\nop{in that} human evaluators may then be able to give more informative feedback than real users.

To further reduce\nop{truly bridge} the discrepancy {between human evaluation and real use cases}\nop{in evaluating semantic parsers with humans}, {one possible solution is to ask human evaluators to come up with questions from scratch (instead of using pre-existing test questions), which guarantees intent understanding. While this solution may still require exposure of table content to evaluators (such that they can have some sense of each table attribute), overexposure can be mitigated by showing them only part (e.g., just a few rows) of the table content, similar to the annotation strategy by \citet{zhong2017seq2sql}. Furthermore, the reduced controllability on the complexity of the evaluator-composed questions can be compensated by conducting human evaluation in a larger scale. We plan to explore this setting in future work.} 

\section{Conclusion and Future Work}
\label{sec:future}
{This work proposes a new and unified framework for the interactive semantic parsing task, named MISP, and instantiates it successfully on the text-to-SQL task.}
We outline several future directions to further improve MISP-SQL and develop MISP systems for other semantic parsing tasks:

\paragraph{Improving Agent Components.}
The flexibility of MISP allows improving on each agent component separately. Take the error detector for example. One can augment the probability-based error detector in MISP-SQL with probability calibration, which has been shown useful in aligning model confidence with its reliability \cite{guo2017calibration}. One can also use learning-based approaches, such as a reinforced decision policy \cite{yao2018interactive}, to increase the rate of identifying wrong and solvable predictions.

\paragraph{Lifelong Learning for Semantic Parsing.}
Learning from user feedback is a promising solution for lifelong semantic parser improvement \cite{iyer2017learning, padmakumar2017integrated, labutov2018learning}. However, this may lead to a non-stationary environment (e.g., changing state transition) from the perspective of the agent, making its training (e.g., error detector learning) unstable. In the context of dialog systems, \citet{padmakumar2017integrated} suggests that this effect can be mitigated by jointly updating the dialog policy and the semantic parser batchwisely. We leave exploring this aspect in our task to future work.



\paragraph{Scaling Up.} It is important for MISP agents to scale to larger backend data sources (e.g., knowledge bases like Freebase or Wikidata). To this end, one can improve MISP from at least three aspects: (1) using more intelligent interaction designs (e.g., free-form text as user feedback) to speed up the hypothesis space searching globally, (2) strengthening the world model to nail down a smaller set of plausible hypotheses based on both the initial question and user feedback, and (3) training the agent to learn to improve the parsing accuracy while minimizing the number of required human interventions over time.

\section*{Acknowledgments}
{This research was sponsored in part by the Army Research Office under cooperative agreements W911NF-17-1-0412, NSF Grant IIS1815674, Fujitsu gift grant, and Ohio Supercomputer
Center \cite{OhioSupercomputerCenter1987}. The views and conclusions contained herein are those of the authors and should not be interpreted as representing the official policies, either expressed or implied, of the Army Research Office or the U.S. Government. The U.S. Government is authorized to reproduce and distribute reprints for Government purposes notwithstanding any copyright notice herein.}

\bibliography{reference}

\begin{thebibliography}{43}
\expandafter\ifx\csname natexlab\endcsname\relax\def\natexlab#1{#1}\fi

\bibitem[{Artzi and Zettlemoyer(2013)}]{artzi2013weakly}
Yoav Artzi and Luke Zettlemoyer. 2013.
\newblock Weakly supervised learning of semantic parsers for mapping
  instructions to actions.
\newblock \emph{Transactions of the Association for Computational Linguistics},
  1:49--62.

\bibitem[{Berant et~al.(2013)Berant, Chou, Frostig, and
  Liang}]{berant2013semantic}
Jonathan Berant, Andrew Chou, Roy Frostig, and Percy Liang. 2013.
\newblock Semantic parsing on freebase from question-answer pairs.
\newblock In \emph{Proceedings of the 2013 Conference on Empirical Methods in
  Natural Language Processing}, pages 1533--1544.

\bibitem[{Berant et~al.(2019)Berant, Deutch, Globerson, Milo, and
  Wolfson}]{berant2019explaining}
Jonathan Berant, Daniel Deutch, Amir Globerson, Tova Milo, and Tomer Wolfson.
  2019.
\newblock Explaining queries over web tables to non-experts.
\newblock In \emph{Proceedings of the 35th IEEE International Conference on
  Data Engineering (ICDE)}.

\bibitem[{Bogin et~al.(2019)Bogin, Gardner, and Berant}]{bogin2019representing}
Ben Bogin, Matt Gardner, and Jonathan Berant. 2019.
\newblock Representing schema structure with graph neural networks for
  text-to-{SQL} parsing.
\newblock In \emph{Proceedings of the 57th Annual Meeting of the Association
  for Computational Linguistics}, pages 4560--4565.

\bibitem[{Campagna et~al.(2017)Campagna, Ramesh, Xu, Fischer, and
  Lam}]{campagna2017almond}
Giovanni Campagna, Rakesh Ramesh, Silei Xu, Michael Fischer, and Monica~S Lam.
  2017.
\newblock Almond: The architecture of an open, crowdsourced,
  privacy-preserving, programmable virtual assistant.
\newblock In \emph{Proceedings of the 26th International Conference on World
  Wide Web}, pages 341--350. International World Wide Web Conferences Steering
  Committee.

\bibitem[{Center(1987)}]{OhioSupercomputerCenter1987}
Ohio~Supercomputer Center. 1987.
\newblock Ohio supercomputer center.
\newblock \url{http://osc.edu/ark:/19495/f5s1ph73}.

\bibitem[{Chaurasia and Mooney(2017)}]{chaurasia2017dialog}
Shobhit Chaurasia and Raymond~J Mooney. 2017.
\newblock Dialog for language to code.
\newblock In \emph{Proceedings of the Eighth International Joint Conference on
  Natural Language Processing (Volume 2: Short Papers)}, pages 175--180.

\bibitem[{Devlin et~al.(2019)Devlin, Chang, Lee, and
  Toutanova}]{devlin2018bert}
Jacob Devlin, Ming-Wei Chang, Kenton Lee, and Kristina Toutanova. 2019.
\newblock {BERT}: Pre-training of deep bidirectional transformers for language
  understanding.
\newblock In \emph{Proceedings of the 2019 Conference of the North American
  Chapter of the Association for Computational Linguistics: Human Language
  Technologies, Volume 1 (Long and Short Papers)}, pages 4171--4186.

\bibitem[{Dong and Lapata(2016)}]{dong2016language}
Li~Dong and Mirella Lapata. 2016.
\newblock Language to logical form with neural attention.
\newblock In \emph{Proceedings of the 54th Annual Meeting of the Association
  for Computational Linguistics (Volume 1: Long Papers)}, volume~1, pages
  33--43.

\bibitem[{Dong et~al.(2018)Dong, Quirk, and Lapata}]{dong2018confidence}
Li~Dong, Chris Quirk, and Mirella Lapata. 2018.
\newblock Confidence modeling for neural semantic parsing.
\newblock In \emph{Proceedings of the 56th Annual Meeting of the Association
  for Computational Linguistics (Volume 1: Long Papers)}, volume~1, pages
  743--753.

\bibitem[{Finegan-Dollak et~al.(2018)Finegan-Dollak, Kummerfeld, Zhang,
  Ramanathan, Sadasivam, Zhang, and Radev}]{finegan2018improving}
Catherine Finegan-Dollak, Jonathan~K Kummerfeld, Li~Zhang, Karthik Ramanathan,
  Sesh Sadasivam, Rui Zhang, and Dragomir Radev. 2018.
\newblock Improving text-to-{SQL} evaluation methodology.
\newblock In \emph{Proceedings of the 56th Annual Meeting of the Association
  for Computational Linguistics (Volume 1: Long Papers)}, pages 351--360.

\bibitem[{Gal and Ghahramani(2016)}]{gal2016dropout}
Yarin Gal and Zoubin Ghahramani. 2016.
\newblock Dropout as a {Bayesian} approximation: Representing model uncertainty
  in deep learning.
\newblock In \emph{international conference on machine learning}, pages
  1050--1059.

\bibitem[{Guo et~al.(2017)Guo, Pleiss, Sun, and
  Weinberger}]{guo2017calibration}
Chuan Guo, Geoff Pleiss, Yu~Sun, and Kilian~Q Weinberger. 2017.
\newblock On calibration of modern neural networks.
\newblock In \emph{Proceedings of the 34th International Conference on Machine
  Learning-Volume 70}, pages 1321--1330. JMLR. org.

\bibitem[{Guo et~al.(2019)Guo, Zhan, Gao, Xiao, Lou, Liu, and
  Zhang}]{guo-etal-2019-towards}
Jiaqi Guo, Zecheng Zhan, Yan Gao, Yan Xiao, Jian-Guang Lou, Ting Liu, and
  Dongmei Zhang. 2019.
\newblock Towards complex text-to-{SQL} in cross-domain database with
  intermediate representation.
\newblock In \emph{Proceedings of the 57th Annual Meeting of the Association
  for Computational Linguistics}, pages 4524--4535.

\bibitem[{Gur et~al.(2018)Gur, Yavuz, Su, and Yan}]{gur2018dialsql}
Izzeddin Gur, Semih Yavuz, Yu~Su, and Xifeng Yan. 2018.
\newblock {DialSQL}: Dialogue based structured query generation.
\newblock In \emph{Proceedings of the 56th Annual Meeting of the Association
  for Computational Linguistics (Volume 1: Long Papers)}, pages 1339--1349.

\bibitem[{Ha and Schmidhuber(2018)}]{ha2018world}
David Ha and J{\"u}rgen Schmidhuber. 2018.
\newblock World models.
\newblock \emph{ArXiv preprint arXiv:1803.10122}.

\bibitem[{He et~al.(2016)He, Michael, Lewis, and Zettlemoyer}]{he2016human}
Luheng He, Julian Michael, Mike Lewis, and Luke Zettlemoyer. 2016.
\newblock Human-in-the-loop parsing.
\newblock In \emph{Proceedings of the 2016 Conference on Empirical Methods in
  Natural Language Processing}, pages 2337--2342.

\bibitem[{Hendrycks and Gimpel(2017)}]{hendrycks17baseline}
Dan Hendrycks and Kevin Gimpel. 2017.
\newblock A baseline for detecting misclassified and out-of-distribution
  examples in neural networks.
\newblock In \emph{Proceedings of International Conference on Learning
  Representations}.

\bibitem[{Hwang et~al.(2019)Hwang, Yim, Park, and Seo}]{hwang2019comprehensive}
Wonseok Hwang, Jinyeung Yim, Seunghyun Park, and Minjoon Seo. 2019.
\newblock A comprehensive exploration on {WikiSQL} with table-aware word
  contextualization.
\newblock \emph{ArXiv preprint arXiv:1902.01069}.

\bibitem[{Iyer et~al.(2017)Iyer, Konstas, Cheung, Krishnamurthy, and
  Zettlemoyer}]{iyer2017learning}
Srinivasan Iyer, Ioannis Konstas, Alvin Cheung, Jayant Krishnamurthy, and Luke
  Zettlemoyer. 2017.
\newblock Learning a neural semantic parser from user feedback.
\newblock In \emph{Proceedings of the 55th Annual Meeting of the Association
  for Computational Linguistics (Volume 1: Long Papers)}, pages 963--973.

\bibitem[{Iyer et~al.(2016)Iyer, Konstas, Cheung, and
  Zettlemoyer}]{iyer2016summarizing}
Srinivasan Iyer, Ioannis Konstas, Alvin Cheung, and Luke Zettlemoyer. 2016.
\newblock Summarizing source code using a neural attention model.
\newblock In \emph{Proceedings of the 54th Annual Meeting of the Association
  for Computational Linguistics (Volume 1: Long Papers)}, volume~1, pages
  2073--2083.

\bibitem[{Koutrika et~al.(2010)Koutrika, Simitsis, and
  Ioannidis}]{koutrika2010explaining}
Georgia Koutrika, Alkis Simitsis, and Yannis~E Ioannidis. 2010.
\newblock Explaining structured queries in natural language.
\newblock In \emph{2010 IEEE 26th International Conference on Data Engineering
  (ICDE 2010)}, pages 333--344. IEEE.

\bibitem[{Labutov et~al.(2018)Labutov, Yang, and
  Mitchell}]{labutov2018learning}
Igor Labutov, Bishan Yang, and Tom Mitchell. 2018.
\newblock Learning to learn semantic parsers from natural language supervision.
\newblock In \emph{Proceedings of the 2018 Conference on Empirical Methods in
  Natural Language Processing}, pages 1676--1690.

\bibitem[{Li and Jagadish(2014)}]{li2014constructing}
Fei Li and HV~Jagadish. 2014.
\newblock Constructing an interactive natural language interface for relational
  databases.
\newblock \emph{Proceedings of the VLDB Endowment}, 8(1):73--84.

\bibitem[{Ngonga~Ngomo et~al.(2013)Ngonga~Ngomo, B{\"u}hmann, Unger, Lehmann,
  and Gerber}]{ngonga2013sorry}
Axel-Cyrille Ngonga~Ngomo, Lorenz B{\"u}hmann, Christina Unger, Jens Lehmann,
  and Daniel Gerber. 2013.
\newblock Sorry, {I} don't speak {SPARQL}: translating {SPARQL} queries into
  natural language.
\newblock In \emph{Proceedings of the 22nd international conference on World
  Wide Web}, pages 977--988. ACM.

\bibitem[{Padmakumar et~al.(2017)Padmakumar, Thomason, and
  Mooney}]{padmakumar2017integrated}
Aishwarya Padmakumar, Jesse Thomason, and Raymond~J Mooney. 2017.
\newblock Integrated learning of dialog strategies and semantic parsing.
\newblock In \emph{Proceedings of the 15th Conference of the European Chapter
  of the Association for Computational Linguistics: Volume 1, Long Papers},
  pages 547--557.

\bibitem[{Quirk et~al.(2015)Quirk, Mooney, and Galley}]{quirk2015language}
Chris Quirk, Raymond Mooney, and Michel Galley. 2015.
\newblock Language to code: Learning semantic parsers for {If-This-Then-That}
  recipes.
\newblock In \emph{Proceedings of the 53rd Annual Meeting of the Association
  for Computational Linguistics and the 7th International Joint Conference on
  Natural Language Processing (Volume 1: Long Papers)}, volume~1, pages
  878--888.

\bibitem[{Russell and Norvig(2009)}]{russell2009artificial}
Stuart Russell and Peter Norvig. 2009.
\newblock Artificial intelligence: A modern approach.

\bibitem[{Siddhant and Lipton(2018)}]{siddhant2018deep}
Aditya Siddhant and Zachary~C Lipton. 2018.
\newblock Deep {Bayesian} active learning for natural language processing:
  Results of a large-scale empirical study.
\newblock In \emph{Proceedings of the 2018 Conference on Empirical Methods in
  Natural Language Processing}, pages 2904--2909.

\bibitem[{Srivastava et~al.(2014)Srivastava, Hinton, Krizhevsky, Sutskever, and
  Salakhutdinov}]{srivastava2014dropout}
Nitish Srivastava, Geoffrey Hinton, Alex Krizhevsky, Ilya Sutskever, and Ruslan
  Salakhutdinov. 2014.
\newblock Dropout: a simple way to prevent neural networks from overfitting.
\newblock \emph{The Journal of Machine Learning Research}, 15(1):1929--1958.

\bibitem[{Su et~al.(2017)Su, Awadallah, Khabsa, Pantel, Gamon, and
  Encarnacion}]{su2017building}
Yu~Su, Ahmed~Hassan Awadallah, Madian Khabsa, Patrick Pantel, Michael Gamon,
  and Mark Encarnacion. 2017.
\newblock Building natural language interfaces to {Web API}s.
\newblock In \emph{Proceedings of the 2017 ACM on Conference on Information and
  Knowledge Management}, pages 177--186. ACM.

\bibitem[{Su et~al.(2018)Su, Awadallah, Wang, and White}]{su2018natural}
Yu~Su, Ahmed~Hassan Awadallah, Miaosen Wang, and Ryen~W White. 2018.
\newblock Natural language interfaces with fine-grained user interaction: A
  case study on web {API}s.
\newblock In \emph{Proceedings of the International ACM SIGIR Conference on
  Research and Development in Information Retrieval}.

\bibitem[{Thomason et~al.(2015)Thomason, Zhang, Mooney, and
  Stone}]{thomason2015learning}
Jesse Thomason, Shiqi Zhang, Raymond~J Mooney, and Peter Stone. 2015.
\newblock Learning to interpret natural language commands through human-robot
  dialog.
\newblock In \emph{Twenty-Fourth International Joint Conference on Artificial
  Intelligence}.

\bibitem[{Wang et~al.(2018)Wang, Brockschmidt, and Singh}]{wang2018pointing}
Chenglong Wang, Marc Brockschmidt, and Rishabh Singh. 2018.
\newblock Pointing out {SQL} queries from text.

\bibitem[{Wang et~al.(2015)Wang, Berant, and Liang}]{wang2015building}
Yushi Wang, Jonathan Berant, and Percy Liang. 2015.
\newblock Building a semantic parser overnight.
\newblock In \emph{Proceedings of the 53rd Annual Meeting of the Association
  for Computational Linguistics and the 7th International Joint Conference on
  Natural Language Processing (Volume 1: Long Papers)}, volume~1, pages
  1332--1342.

\bibitem[{Xiao and Wang(2019)}]{xiao2018quantifying}
Yijun Xiao and William~Yang Wang. 2019.
\newblock Quantifying uncertainties in natural language processing tasks.
\newblock In \emph{Proceedings of the AAAI Conference on Artificial
  Intelligence}, volume~33, pages 7322--7329.

\bibitem[{Xu et~al.(2018)Xu, Wu, Wang, Feng, and Sheinin}]{xu2018sql}
Kun Xu, Lingfei Wu, Zhiguo Wang, Yansong Feng, and Vadim Sheinin. 2018.
\newblock {SQL}-to-text generation with graph-to-sequence model.
\newblock In \emph{Proceedings of the 2018 Conference on Empirical Methods in
  Natural Language Processing}, pages 931--936.

\bibitem[{Xu et~al.(2017)Xu, Liu, and Song}]{xu2017sqlnet}
Xiaojun Xu, Chang Liu, and Dawn Song. 2017.
\newblock {SQLNet}: Generating structured queries from natural language without
  reinforcement learning.
\newblock \emph{ArXiv preprint arXiv:1711.04436}.

\bibitem[{Yao et~al.(2019)Yao, Li, Gao, Sadler, and Sun}]{yao2018interactive}
Ziyu Yao, Xiujun Li, Jianfeng Gao, Brian Sadler, and Huan Sun. 2019.
\newblock Interactive semantic parsing for if-then recipes via hierarchical
  reinforcement learning.
\newblock In \emph{Proceedings of the AAAI Conference on Artificial
  Intelligence}, volume~33, pages 2547--2554.

\bibitem[{Yu et~al.(2018{\natexlab{a}})Yu, Li, Zhang, Zhang, and
  Radev}]{yu2018typesql}
Tao Yu, Zifan Li, Zilin Zhang, Rui Zhang, and Dragomir Radev.
  2018{\natexlab{a}}.
\newblock {TypeSQL}: Knowledge-based type-aware neural text-to-{SQL}
  generation.
\newblock In \emph{Proceedings of the 2018 Conference of the North American
  Chapter of the Association for Computational Linguistics: Human Language
  Technologies, Volume 2 (Short Papers)}, pages 588--594.

\bibitem[{Yu et~al.(2018{\natexlab{b}})Yu, Yasunaga, Yang, Zhang, Wang, Li, and
  Radev}]{yu2018syntaxsqlnet}
Tao Yu, Michihiro Yasunaga, Kai Yang, Rui Zhang, Dongxu Wang, Zifan Li, and
  Dragomir Radev. 2018{\natexlab{b}}.
\newblock {SyntaxSQLNet}: Syntax tree networks for complex and cross-domain
  text-to-{SQL} task.
\newblock In \emph{Proceedings of the 2018 Conference on Empirical Methods in
  Natural Language Processing}, pages 1653--1663.

\bibitem[{Yu et~al.(2018{\natexlab{c}})Yu, Zhang, Yang, Yasunaga, Wang, Li, Ma,
  Li, Yao, Roman et~al.}]{yu2018spider}
Tao Yu, Rui Zhang, Kai Yang, Michihiro Yasunaga, Dongxu Wang, Zifan Li, James
  Ma, Irene Li, Qingning Yao, Shanelle Roman, et~al. 2018{\natexlab{c}}.
\newblock Spider: A large-scale human-labeled dataset for complex and
  cross-domain semantic parsing and text-to-{SQL} task.
\newblock In \emph{Proceedings of the 2018 Conference on Empirical Methods in
  Natural Language Processing}, pages 3911--3921.

\bibitem[{Zhong et~al.(2017)Zhong, Xiong, and Socher}]{zhong2017seq2sql}
Victor Zhong, Caiming Xiong, and Richard Socher. 2017.
\newblock Seq2{SQL}: Generating structured queries from natural language using
  reinforcement learning.
\newblock \emph{ArXiv preprint arXiv:1709.00103}.

\end{thebibliography}
\bibliographystyle{acl_natbib}

\newpage
\appendix
\clearpage

\section{Extension to Complex SQL} \label{app:ext_spider}
Table~\ref{tab:spider_grammar} shows the extended lexicon entries and grammar rules in NLG for applying our MISP-SQL agent to generate more complex SQL queries, such as those on Spider \cite{yu2018spider}. In this dataset, a SQL query can associate with multiple tables. Therefore, we name a column by combining the column name with its table name (i.e., ``$col$'' in table ``$tab$'' $\rightarrow$ \textsc{Col}[$col$ (table $tab$)]). For simplicity, we omit ``(table $tab$)'' when referring to a column $col$ in the grammar.

\section{Simulation Evaluation Results} \label{app:sim_results}
The complete simulation experiment results of MISP-SQL agents (based on SQLNet and SQLova) are shown in Table~\ref{tab:complete_sqlnet} \& \ref{tab:complete_sqlova}.

\section{Error Detector Comparison}\label{app:error_detector}
As a supplementary experiment to Figure~\ref{fig:sim_results}, in this section, we show the performance of different error detectors under the same average number of questions (``\textit{target budget}''). Specifically, for each base semantic parser and each kind of error detector, we tune its decision threshold (i.e., $p^*$ and $s^*$) such that the resulting average number of questions (``\textit{actual budget}'') is as close to the target as possible. In practice, we relax the actual budget to be within $\pm 0.015$ of the target budget, which empirically leads to merely negligible variance. The results are shown in Table~\ref{tab:budget_sqlnet_dev}-\ref{tab:budget_sqlnet_test} for SQLNet and Table~\ref{tab:budget_sqlova_dev}-\ref{tab:budget_sqlova_test} for SQLova.

\newpage
\begin{table}[ht]
    \centering
    \begin{tabular}{l c c c}\toprule
    & \multicolumn{3}{c}{\textbf{SQLNet}} \\
        \textbf{System} & \textbf{Acc\textsubscript{qm}} & \textbf{Acc\textsubscript{ex}} & \textbf{Avg. \#q}\\\midrule
        no interaction & 0.615 & 0.681 & N/A  \\\midrule
        MISP-SQL\textsuperscript{Unlimit10} & 0.932 & 0.948 & 7.445 \\
        MISP-SQL\textsuperscript{Unlimit3} & 0.870 & 0.900 & 7.052 \\\midrule
        MISP-SQL\textsuperscript{$p^*$=0.95} & 0.782 & 0.824 & 1.713 \\
        MISP-SQL\textsuperscript{$p^*$=0.8} & 0.729 & 0.779 & 1.104 \\
        MISP-SQL\textsuperscript{$p^*$=0.5} & 0.661 & 0.722 & 0.421 \\
        MISP-SQL\textsuperscript{$s^*$=0.01} & 0.796 & 0.845 & 2.106 \\
        MISP-SQL\textsuperscript{$s^*$=0.05} & 0.725 & 0.786 & 1.348 \\
        MISP-SQL\textsuperscript{$s^*$=0.1} & 0.695 & 0.758 & 1.009 \\
        MISP-SQL\textsuperscript{$s^*$=0.2} & 0.650 & 0.714 & 0.413 \\
        \bottomrule
    \end{tabular}
    \caption{Simulation evaluation of MISP-SQL (based on SQLNet) on WikiSQL Test set.}
    \label{tab:complete_sqlnet}
\end{table}

\begin{table}[ht]
    \centering
    \begin{tabular}{l c c c }\toprule
         & \multicolumn{3}{c}{\textbf{SQLova}} \\
        \textbf{System} & \textbf{Acc\textsubscript{qm}} & \textbf{Acc\textsubscript{ex}} & \textbf{Avg. \#q}\\\midrule
        no interaction & 0.797 & 0.853 & N/A  \\\midrule
        MISP-SQL\textsuperscript{Unlimit10} & 0.985 & 0.991 & 6.591 \\
        MISP-SQL\textsuperscript{Unlimit3} & 0.955 & 0.974 & 6.515 \\\midrule
        MISP-SQL\textsuperscript{$p^*$=0.95} & 0.912 & 0.939 & 0.773 \\
        MISP-SQL\textsuperscript{$p^*$=0.8} & 0.880 & 0.914 & 0.488 \\
        MISP-SQL\textsuperscript{$p^*$=0.5} & 0.835 & 0.879 & 0.209 \\
        MISP-SQL\textsuperscript{$s^*$=0.01} & 0.913 & 0.942 &0.893 \\
        MISP-SQL\textsuperscript{$s^*$=0.03} & 0.866 & 0.912 & 0.515 \\
        MISP-SQL\textsuperscript{$s^*$=0.05} & 0.840 & 0.892 & 0.333 \\
        MISP-SQL\textsuperscript{$s^*$=0.07} & 0.825 & 0.880 & 0.216 \\
        \bottomrule
    \end{tabular}
    \caption{Simulation evaluation of MISP-SQL (based on SQLova) on WikiSQL Test set.}
    \label{tab:complete_sqlova}
\end{table}

\begin{table*}[ht]
    \centering\small
\hspace*{-1cm}\resizebox{1.1\textwidth}{!}{
    \begin{tabular}{lc}\toprule
    \multicolumn{2}{c}{\textbf{[Lexicon]}}\\[1mm]
    \multicolumn{2}{c}{is greater than (or equivalent to)\textbar equals to\textbar is less than (or equivalent to)\textbar does not equal to $\rightarrow$ \textsc{Op}[\texttt{$>$(=)}\textbar \texttt{$=$}\textbar \texttt{$<$(=)}\textbar\texttt{$!=$}]}\\
    \multicolumn{2}{c}{is IN\textbar is NOT IN\textbar follows a pattern like\textbar is between $\rightarrow$ \textsc{Op}[\texttt{in}\textbar\texttt{not} \texttt{in}\textbar\texttt{like}\textbar\texttt{between}]}\\
    \multicolumn{2}{c}{sum of values in\textbar average value in\textbar number of\textbar minimum value in\textbar maximum value in $\rightarrow$ \textsc{Agg}[\texttt{sum}\textbar\texttt{avg}\textbar\texttt{count}\textbar\texttt{min}\textbar\texttt{max}]} \\
    \multicolumn{2}{c}{in descending order (and limited to top N)\textbar in ascending order (and limited to top N) $\rightarrow$ \textsc{Order}[\texttt{desc(limit N)}\textbar \texttt{asc(limit N)}]}\\
    \midrule
    \multicolumn{2}{c}{\textbf{[Grammar]}}\\[1mm]
    (R1) & ``$col$'' in table ``$tab$'' $\rightarrow$ \textsc{Col}[$col$ (table $tab$)] \\
    (R2) & Does the system need to return information about \textsc{Col}[$col$] ? $\rightarrow$ Q[$col\|$\texttt{SELECT} $agg?$ $col$]\\
    (R3) & Does the system need to return \textsc{Agg}[$agg$] \textsc{Col}[$col$] ? $\rightarrow$ Q[$agg\|$\texttt{SELECT} $agg$ $col$] \\
    (R4) & Does the system need to return a value \underline{after} any mathematical calculations on \textsc{Col}[$col$] ? $\rightarrow$ Q[$agg$=None$\|$\texttt{SELECT} $agg$ $col$] \\
    (R5) & Does the system need to consider any conditions about \textsc{Col}[$col$] ? $\rightarrow$ Q[$col\|$\texttt{WHERE} $col$ $op$ $val$] \\
    (R6) & The system considers the following condition: \textsc{Col}[$col$] \textsc{Op}[$op$] a given literal value. Is this condition correct? $\rightarrow$\\ & Q[$terminal\|$\texttt{WHERE} $col$ $op$ $terminal$] \\
    (R7) & The system considers the following condition: \textsc{Col}[$col$] \textsc{Op}[$op$] a value to be calculated. Is this condition correct? $\rightarrow$\\ & Q[$root\|$\texttt{WHERE} $col$ $op$ $root$] \\
    (R8)& Do the conditions about \textsc{Col}[$col_i$] and \textsc{Col}[$col_j$] hold at the same time? $\rightarrow$ Q[\texttt{AND}$\|$\texttt{WHERE} $col_i$ .. \texttt{AND} $col_j$ ..] \\
    (R9)& Do the conditions about \textsc{Col}[$col_i$] and \textsc{Col}[$col_j$] hold alternatively? $\rightarrow$ Q[\texttt{OR}$\|$\texttt{WHERE} $col_i$ .. \texttt{OR} $col_j$ ..] \\
    (R10) & Does the system need to group items in table $tab$ based on \textsc{Col}[$col$] before doing any mathematical calculations? $\rightarrow$\\ & Q[$col\|$\texttt{GROUP BY} $col$] \\
    (R11)& Given that the system groups items in table $tab^g$ based on \textsc{Col}[$col^g$] before doing any mathematical calculations,\\& does the system need to consider any conditions about \textsc{Col}[$col$] ? $\rightarrow$ Q[$col\|$\texttt{GROUP BY} $col^g$ \texttt{HAVING} $agg?$ $col$] \\
    (R12)& Given that the system groups items in table $tab^g$ based on \textsc{Col}[$col^g$] before doing any mathematical calculations,\\& does the system need to consider any conditions about \textsc{Agg}[$agg$] \textsc{Col}[$col$] ? $\rightarrow$ Q[$agg\|$\texttt{GROUP BY} $col^g$ \texttt{HAVING} $agg$ $col$]\\
    (R13)& Given that the system groups items in table $tab^g$ based on \textsc{Col}[$col^g$] before doing any mathematical calculations, does the system need to\\\multicolumn{2}{c}{consider a value \underline{after} any mathematical calculations on \textsc{Col}[$col$] ? $\rightarrow$ Q[$agg$=None$\|$\texttt{GROUP BY} $col^g$ \texttt{HAVING} $agg$ $col$]}\\
    (R14)& The system groups items in table $tab^g$ based on \textsc{Col}[$col^g$] before doing any mathematical calculations, then considers the following\\ \multicolumn{2}{c}{condition: \textsc{Col}[$col$] \textsc{Op}[$op$] a value. Is this condition correct? $\rightarrow$ Q[$op\|$\texttt{GROUP BY} $col^g$ \texttt{HAVING} $agg?$ $col$ $op$ $val$]} \\
    (R15) & Given that the system groups items in table $tab^g$ based on \textsc{Col}[$col^g$] before doing any mathematical calculations, does it need to\\&consider any conditions? $\rightarrow$ Q[\texttt{NONE\_HAVING} $\|$\texttt{GROUP BY} $col^g$ \texttt{NONE\_HAVING}] \\
    (R16) & Does the system need to order results based on \textsc{Col}[$col$] ? $\rightarrow$ Q[$col\|$\texttt{ORDER BY} $agg?$ $col$]\\
    (R17) & Does the system need to order results based on \textsc{Agg}[$agg$] \textsc{Col}[$col$] ? $\rightarrow$ Q[$agg\|$\texttt{ORDER BY} $agg$ $col$]\\
    (R18) & Does the system need to order results based on a value \underline{after} any mathematical calculations on \textsc{Col}[$col$] ? $\rightarrow$\\& Q[$agg$=None$\|$\texttt{ORDER BY} $agg$ $col$]\\
    (R19) & Given that the system orders the results based on (\textsc{Agg}[$agg$]) \textsc{Col}[$col$], does it need to be \textsc{Order}[$od$] ? $\rightarrow$\\ & Q[$od\|$\texttt{ORDER BY} $agg?$ $col$ $od$] \\
    \bottomrule
    \end{tabular}}
    \caption{Extended lexicon and grammar for MISP-SQL NLG module to handle complex SQL on Spider.}
    \label{tab:spider_grammar}
\end{table*}

\begin{table}[ht]
    \centering
    \begin{tabular}{ccccc}\toprule
    \textbf{Avg. \#q} & \multicolumn{2}{c}{\textbf{Probability-based}} & \multicolumn{2}{c}{\textbf{Dropout-based}}\\
    & \textbf{Acc\textsubscript{qm}} & \textbf{Acc\textsubscript{ex}} & \textbf{Acc\textsubscript{qm}} & \textbf{Acc\textsubscript{ex}}\\\midrule
        0.5 & 0.672 & 0.732 & 0.663 & 0.726 \\
        1.0 & 0.725 & 0.775 & 0.706 & 0.765 \\
        1.5 & 0.778 & 0.820 & 0.749 & 0.809 \\
        2.0 & 0.812 & 0.848 & 0.796 & 0.845 \\\bottomrule
    \end{tabular}
    \caption{Comparison of error detectors for SQLNet with a target average number of questions on WikiSQL Dev set.}
    \label{tab:budget_sqlnet_dev}
\end{table}

\begin{table}[ht]
    \centering
    \begin{tabular}{ccccc}\toprule
    \textbf{Avg. \#q} & \multicolumn{2}{c}{\textbf{Probability-based}} & \multicolumn{2}{c}{\textbf{Dropout-based}}\\
    & \textbf{Acc\textsubscript{qm}} & \textbf{Acc\textsubscript{ex}} & \textbf{Acc\textsubscript{qm}} & \textbf{Acc\textsubscript{ex}}\\\midrule
        0.5 & 0.669 & 0.729 & 0.656 & 0.720  \\
        1.0 & 0.722 & 0.773 & 0.695 & 0.758  \\
        1.5 & 0.765 & 0.810 & 0.740 & 0.801  \\
        2.0 & 0.805 & 0.844 & 0.790 & 0.842 \\\bottomrule
    \end{tabular}
    \caption{Comparison of error detectors for SQLNet with a target average number of questions on WikiSQL Test set.}
    \label{tab:budget_sqlnet_test}
\end{table}

\begin{table}[ht]
    \centering
    \begin{tabular}{ccccc}\toprule
    \textbf{Avg. \#q} & \multicolumn{2}{c}{\textbf{Probability-based}} & \multicolumn{2}{c}{\textbf{Dropout-based}}\\
    & \textbf{Acc\textsubscript{qm}} & \textbf{Acc\textsubscript{ex}} & \textbf{Acc\textsubscript{qm}} & \textbf{Acc\textsubscript{ex}}\\\midrule
        0.2 & 0.844 & 0.885 & 0.829 & 0.881 \\
        0.4 & 0.876 & 0.910 & 0.856 & 0.905 \\
        0.6 & 0.902 & 0.932 & 0.887 & 0.927 \\
        0.8 & 0.921 & 0.947 & 0.913 & 0.941 \\\bottomrule
    \end{tabular}
    \caption{Comparison of error detectors for SQLova with a target average number of questions on WikiSQL Dev set.}
    \label{tab:budget_sqlova_dev}
\end{table}

\begin{table}[ht]
    \centering
    \begin{tabular}{ccccc}\toprule
    \textbf{Avg. \#q} & \multicolumn{2}{c}{\textbf{Probability-based}} & \multicolumn{2}{c}{\textbf{Dropout-based}}\\
    & \textbf{Acc\textsubscript{qm}} & \textbf{Acc\textsubscript{ex}} & \textbf{Acc\textsubscript{qm}} & \textbf{Acc\textsubscript{ex}}\\\midrule
        0.2 & 0.832 & 0.877 & 0.823 & 0.878 \\
        0.4 & 0.865 & 0.902 & 0.851 & 0.901 \\
        0.6 & 0.895 & 0.926 & 0.881 & 0.922 \\
        0.8 & 0.915 & 0.941 & 0.904 & 0.936\\\bottomrule
    \end{tabular}
    \caption{Comparison of error detectors for SQLova with a target average number of questions on WikiSQL Test set.}
    \label{tab:budget_sqlova_test}
\end{table}

\end{document}